\documentclass{article}

\usepackage{arxiv}

\usepackage[utf8]{inputenc}
\usepackage[T1]{fontenc}
\usepackage{lineno}
\usepackage{subcaption}
\usepackage{pifont}
\usepackage{tikz,float}
\usepackage{tabularx}
\usepackage{url}
\usepackage{xparse}
\usepackage{multirow}
\usepackage{graphicx}
\usepackage{booktabs}
\usepackage[normalem]{ulem}
\useunder{\uline}{\ul}{}
\usepackage[most]{tcolorbox}
\usepackage{xcolor, colortbl}
 \usepackage[hidelinks]{hyperref}
\usetikzlibrary{spy,calc}

\definecolor{InP}{rgb}{0.86, 0.371, 0.34}
\definecolor{Loh}{rgb}{0.569, 0.86, 0.34}
\definecolor{Man}{rgb}{0, 0, 1}
\definecolor{Unet}{rgb}{0.631, 0.34, 0.86}
\definecolor{Swin}{rgb}{0.34, 0.829, 0.86}

\NewDocumentCommand{\statsquare}{ O{#2} m }{%
    (\begin{tikzpicture}
    \fill[#2] (0,0) rectangle (1.8ex, .8ex); 
    \fill[#1] (0,0) -- (1.8ex,0) -- (1.8ex,.8ex) -- (0,.8ex) -- cycle; 
    \end{tikzpicture})
}

\title{GIRAFE: Glottal Imaging Dataset for Advanced Segmentation, Analysis, and Facilitative Playbacks Evaluation} 

\author{ \href{https://orcid.org/0000-0002-6499-5655}{\includegraphics[scale=0.06]{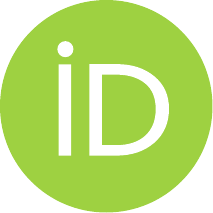}\hspace{1mm}Gustavo Andrade-Miranda}\thanks{Corresponding author.}\\
	Laboratoire de Traitement de l’Information Médicale (LaTIM), \\
	UMR 1101, INSERM\\
        University of Brest\\
	29200, Brest, France. \\
	\texttt{andradema@univ-breast.fr} \\
	\And
	\hspace{1mm}Konstantinos Chatzipapas\\
	Laboratoire de Traitement de l’Information Médicale (LaTIM), \\
	UMR 1101, INSERM\\
        University of Brest\\
	29200, Brest, France. \\
        \\
        3dmi Research Group, Department of Medical Physics\\
        School of Medicine\\
        University of Patras\\
        26504, Rion, Greece. \\
        \And
        \href{https://orcid.org/0000-0002-1928-773X}{\includegraphics[scale=0.06]{orcid.pdf}\hspace{1mm}Julián D. Arias-Londoño}\\
	Department of Signals, Systems, and Radiocommunications\\
	Escuela Técnica Superior de Ingenieros de Telecomunicació\\
        Universidad Politécnica de Madrid\\
	Av. Complutense, 30, 28040, Madrid, Spain. \\
	\texttt{julian.arias@upm.es} \\
        \And
        \href{https://orcid.org/0000-0001-7348-3291}{\includegraphics[scale=0.06]{orcid.pdf}\hspace{1mm}Juan I. Godino-Llorente}\thanks{Corresponding author.}\\
	Department of Signals, Systems, and Radiocommunications\\
	Escuela Técnica Superior de Ingenieros de Telecomunicació\\
        Universidad Politécnica de Madrid\\
	Av. Complutense, 30, 28040, Madrid, Spain. \\
	\texttt{ignacio.godino@upm.es} \\
}

\renewcommand{\shorttitle}{GIRAFE: Glottal Imaging Repository for Advanced Segmentation, Analysis, and Evaluation}

\hypersetup{
pdftitle={A template for the arxiv style},
pdfsubject={q-bio.NC, q-bio.QM},
pdfauthor={David S.~Hippocampus, Elias D.~Striatum},
pdfkeywords={First keyword, Second keyword, More},
}

\begin{document}
\maketitle
\begin{abstract}
The advances in the development of Facilitative Playbacks extracted from High-Speed videoendoscopic sequences of the vocal folds are hindered by a notable lack of publicly available datasets annotated with the semantic segmentations corresponding to the area of the glottal gap. This fact also limits the reproducibility and further exploration of existing research in this field.

To address this gap, GIRAFE is a data repository designed to facilitate the development of advanced techniques for the semantic segmentation, analysis, and fast evaluation of High-Speed videoendoscopic sequences of the vocal folds. The repository includes 65 high-speed videoendoscopic recordings from a cohort of 50 patients (30 female, 20 male). The dataset comprises 15 recordings from healthy controls, 26 from patients with diagnosed voice disorders, and 24 with an unknown health condition. All of them were manually annotated by an expert, including the masks corresponding to the semantic segmentation of the glottal gap. The repository is also complemented with the automatic segmentation of the glottal area using different state-of-the-art approaches. 

This data set has already supported several studies, which demonstrates its usefulness for the development of new glottal gap segmentation algorithms from High-Speed-Videoendoscopic sequences to improve or create new Facilitative Playbacks. Despite these advances and others in the field, the broader challenge of performing an accurate and completely automatic semantic segmentation method of the glottal area remains open.
\end{abstract}

\section{Background}

Voice is fundamental to our daily interactions, serving as the primary means of communication and enabling us to express emotions, thoughts, and cultural identities \cite{Andrade-Miranda2020}. For professionals such as teachers, singers, and public speakers, maintaining a healthy voice is essential for their careers \cite{GODINOLLORENTE2006276,leppavuori2019characterizing,Beaud2022,henrichbernardoni2022}. On average, a person speaks approximately 16,000 words daily, highlighting the extensive use of our vocal folds \cite{Mehl2007}. However, excessive or inappropriate use can lead to voice disorders, such as vocal folds nodules, polyps, laryngitis, or vocal fold paralysis, affecting the quality of life and professional performance \cite{Titze1993,Aronson2009,laver2009}. In this context, clinical voice assessment tools are vital for diagnosing and evaluating these disorders, as they provide a comprehensive and objective analysis of vocal function through techniques such as laryngeal endoscopic imaging, acoustic measurements, and aerodynamic evaluations \cite{andrade2017analyzing}. Among these, laryngeal endoscopic imaging is particularly relevant to distinguish between organic and functional voice disorders, underscoring the need to visually examine the vibratory characteristics of the vocal folds to ensure an accurate diagnosis \cite{svec2007,Shaw2008,Voigt2010b,Pinheiro2012,YAMAUCHI2016205,CROCKER2024}. 

\begin{figure}[ht]
\centering
\includegraphics[width=\textwidth]{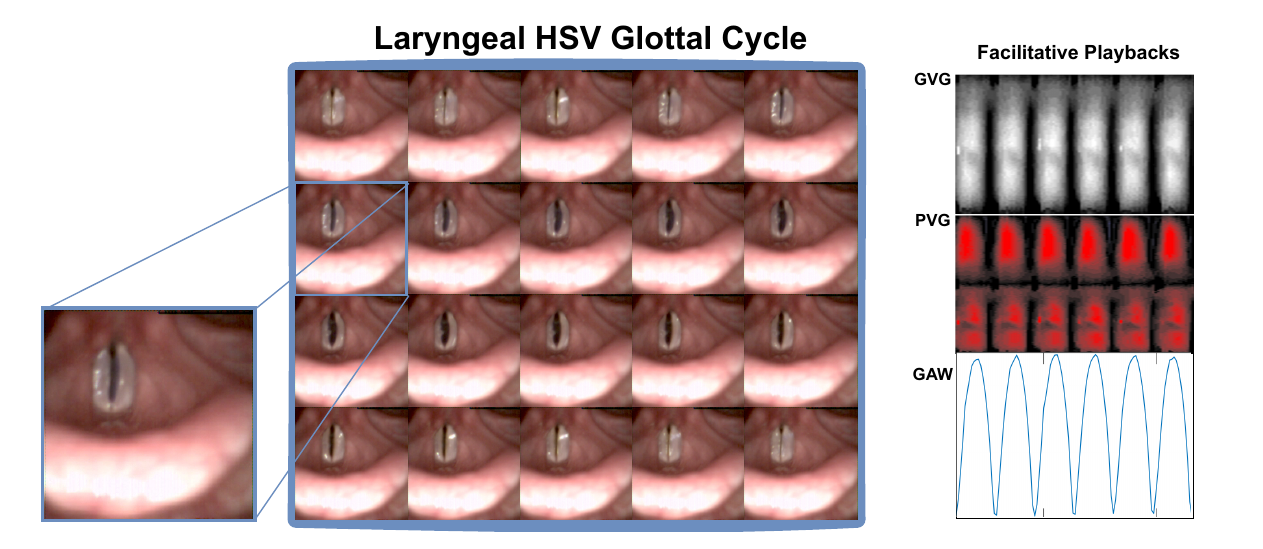}
\caption{Illustration of a complete glottal cycle extracted from laryngeal HSV, along with three FP synthesized from them: GVG, PVG and GAW.}
\label{fig:datasetHSV_FP}
\end{figure}

Among the techniques based on endoscopic imaging, laryngeal high-speed videoendoscopy (HSV) -with sampling frequencies up to 4,000 fps-, is an imaging technique that has shown to be of clinical interest due to its ability to accurately capture the fast dynamics of the vocal folds motion. However, despite its reliability, this technique is not exempt from drawbacks, since it generates a vast amount of data, requires specific and costly hardware, and additional methods to postprocess the HSV sequences recorded. This fact has led to the development of a set of derived techniques that, applied to the HSV  sequences, are aimed to synthesize and represent certain characteristics of the phonation process in a more manageable single image and/or time-varying sequence. These techniques are called Facilitative Playbacks (FP) \cite{Andrade-Miranda2015b,Andrade-Miranda2015c,ANDRADEMIRANDA2017}, and their synthesis is usually based 
on a previous automatic or semi-automatic segmentation of the glottal area \cite{Lohscheller2007,Andrade-Miranda2015,Andrade-Miranda2017Inpa}. Examples of different FP are: glottal area waveforms (GAW) \cite{Andrade-Miranda2020}, glottovibrograms (GVG) \cite{Karakozoglou2012},  phonovibrograms (PVG) \cite{Lohscheller2008b}, and digital kymograms (DKG) \cite{svec1996,svec2002}. The utility of these FP is widely reported in the literature, being essential to evaluate certain aspects of the phonation, including, among others, the periodicity and amplitude of vibration, the presence and characteristics of the mucosal wave (MW), the characteristics of the glottal closure, the duration of the closed phase, the analysis of the symmetry of the vibration pattern, and the identification of non-vibrating instants \cite{andrade2017analyzing,Andrade-Miranda2020}. Figure \ref{fig:datasetHSV_FP} illustrates a complete glottal cycle from an HSV sequence taken from the GIRAFE dataset, along with several FP obtained through automatic segmentation, as described in \cite{Andrade-Miranda2017Inpa}.

Based on an analysis of HSV video sequences and more specifically of certain FP extracted from them, previous studies have shown that the presence of asymmetries, transients, breaks, and irregularities in the vibration of the vocal folds is indicative of the presence of certain phonatory diseases \cite{Wurzbacher2006,Lohscheller2008a,Mehta2013,lohscheller2013,Herbst2014}. For example, the authors in \cite{Yan2007, Ahmad2012, PATEL2017} studied several representations of the GAW, identifying a set of novel quantitative metrics to assess the regularity of the vibration of the vocal folds during phonation. Other studies analyzed acoustic signals and GAW features to establish their mathematical relationships, aiming to identify the parameters that most effectively described the correlation between the characteristics of the GAW and the corresponding acoustic signals \cite{Schelegel2018}.

Similarly, an atypical magnitude and asymmetry of the MW during the vocal folds vibration have been linked to the presence of voice disorders \cite{Shaw2008, Voigt2010b, Krausert2011}. The MW visualized in the HSV sequences, and more specifically using the FP, has been used to diagnose vocal folds lesions and to evaluate treatment outcomes, including the effects of hydration and mucosal healing after phonosurgery \cite{Kaneko2017, Li2015}. Furthermore, the appearance of the MW is recognized as a key component in the myoelastic-aerodynamic theory of phonation, which explains the mechanisms underlying self-sustained vocal fold oscillations \cite{Herbst2013, Elemans2015, Herbst2016a}.

Laryngeal HSV sequences have also been used to investigate functional voice disorders, accurately distinguishing normal and abnormal vibration patterns of the vocal folds based on features derived from different FP \cite{Voigt2010, Schlegel2020}. In this regard, a computer-aided approach was developed to classify vocal folds vibrations from features extracted using image-processing techniques. A classifier was used to separate between normal and pathological vibrations with high accuracy \cite{Voigt2010b}. In addition, laryngeal HSV sequences and their corresponding FP have been applied to detect vocal folds cancer in the early stages. In this regard, a computerized system was used to differentiate malignant from precancerous lesions, showing that precancerous lesions significantly alter the dynamics of the vocal folds \cite{Unger2015}.

The undisputed usefulness of the FP is obscured by the need for a previous automatic or semiautomatic segmentation of the glottal area \cite{Andrade-Miranda2020}. This process is not always simple and is not free from errors, propagating inaccuracies in the representation of the FP, which is consequent with a further incorrect clinical evaluation of the relevant aspects of the phonation process.  

However, with the advent of deep learning (DL), the segmentation process of the glottal area from laryngeal HSV sequences has been significantly automated \cite{Kist2021,Kist2022,CHEN2023e14242,Pedersen2023} using methods based on convolutional neural networks (CNN) and transformer-based models, eliminating the need for hand-crafted features \cite{Conze2023,ANDRADEMIRANDA2023}. Nevertheless, the performance of the DL models is significantly limited by the amount of annotated data available for training \cite{azad2023foundational,wang2023sammed3d}, since leading approaches are notoriously data-hungry. On the other hand, laryngeal HSV data and its annotations are typically scarce and difficult to obtain due to legal regulations, the rarity of certain diseases, the high cost and complexity of acquiring expert-level annotations \cite{Alzubaidi2023}, and the high cost of the HSV devices used. These reasons explain the lack of public datasets developed for this purpose. 

On the other hand, the training and comparison of different automatic segmentation algorithms of the glottal gap from HSV sequences require the availability of large manually annotated datasets to be used as a gold standard for supervising the process. Besides, different datasets are also required not only for training and validation purposes but also to evaluate the generalization abilities of the methods developed \cite{ANDRADEMIRANDA2024}.

In view of the aforementioned and following the examples in other modalities, such as Magnetic Resonance Imaging (MRI) and Computer Tomography (CT) for full-body organ segmentation \cite{kavur2021chaos,ji2022amos,ma2023unleashingstrengthsunlabeleddata,Koitka2024}, collaborative efforts are needed to provide new datasets with annotated laryngeal HSV sequences. The experience in these other domains demonstrates that the availability of open data has spurred innovation, collaboration, and research advances, leading to the creation of foundational models that were later reused for various downstream tasks \cite{liu2023clip,wang2023sammed3d}. 

To our knowledge, the only publicly available dataset in this context is the BAGLS dataset \cite{Gomez2020}. The dataset comprises 640 recordings, including 380 from healthy individuals and 262 from subjects with diagnosed disorders or noticeably affected vocal fold oscillations. Health status information is unavailable for 50 subjects. The disorders range from functional to organic conditions. Of the total recordings, 432 are from female subjects, 177 from male subjects, and 31 have unspecified gender information. Sampling rates range from 1,000 Hz to 10,000 Hz, with image resolutions between $256 \times 120$ px and $512 \times 512$ px. The HSVs were acquired using five different cameras and three distinct lighting systems, utilizing both rigid oral endoscopes (with 70° and 90° angles) and flexible nasal endoscopes of two diameters. Of the 640 videos, 618 are in grayscale, and 22 are in RGB. 

Despite the extensive corpus and variability offered by the BAGLS dataset, the majority of its videos are in grayscale. In contrast, the GIRAFE dataset offers all recordings in color, providing enhanced visual detail that may improve segmentation accuracy and facilitate the detection of subtle vocal fold abnormalities, such as polyps, cysts, and nodules, which are less discernible in grayscale. Moreover, GIRAFE includes a comprehensive evaluation of FP, which is not available in BAGLS, as well as glottal gap segmentation using both traditional and deep learning-based image processing techniques. Rather than competing with BAGLS, GIRAFE is intended to complement it by enabling the training and evaluation of more robust deep learning (DL) models, as well as the fine-tuning of existing ones. By openly sharing the GIRAFE dataset, we anticipate advancements in FP development with significant clinical implications. Furthermore, GIRAFE enables objective comparisons of automatic segmentation methods and complements BAGLS by providing a comprehensive resource to ensure robustness across diverse datasets. This is expected to contribute to the development of foundational models that are more generalizable and effective in real-world clinical applications \cite{MAZUROWSKI2023102918,Siblini2024SAM}.

GIRAFE comprises recordings from a diverse patient population and encompasses different voice disorders, including annotations meticulously performed and rigorously reviewed. 

\begin{figure}[ht]
\centering
\includegraphics[width=\textwidth]{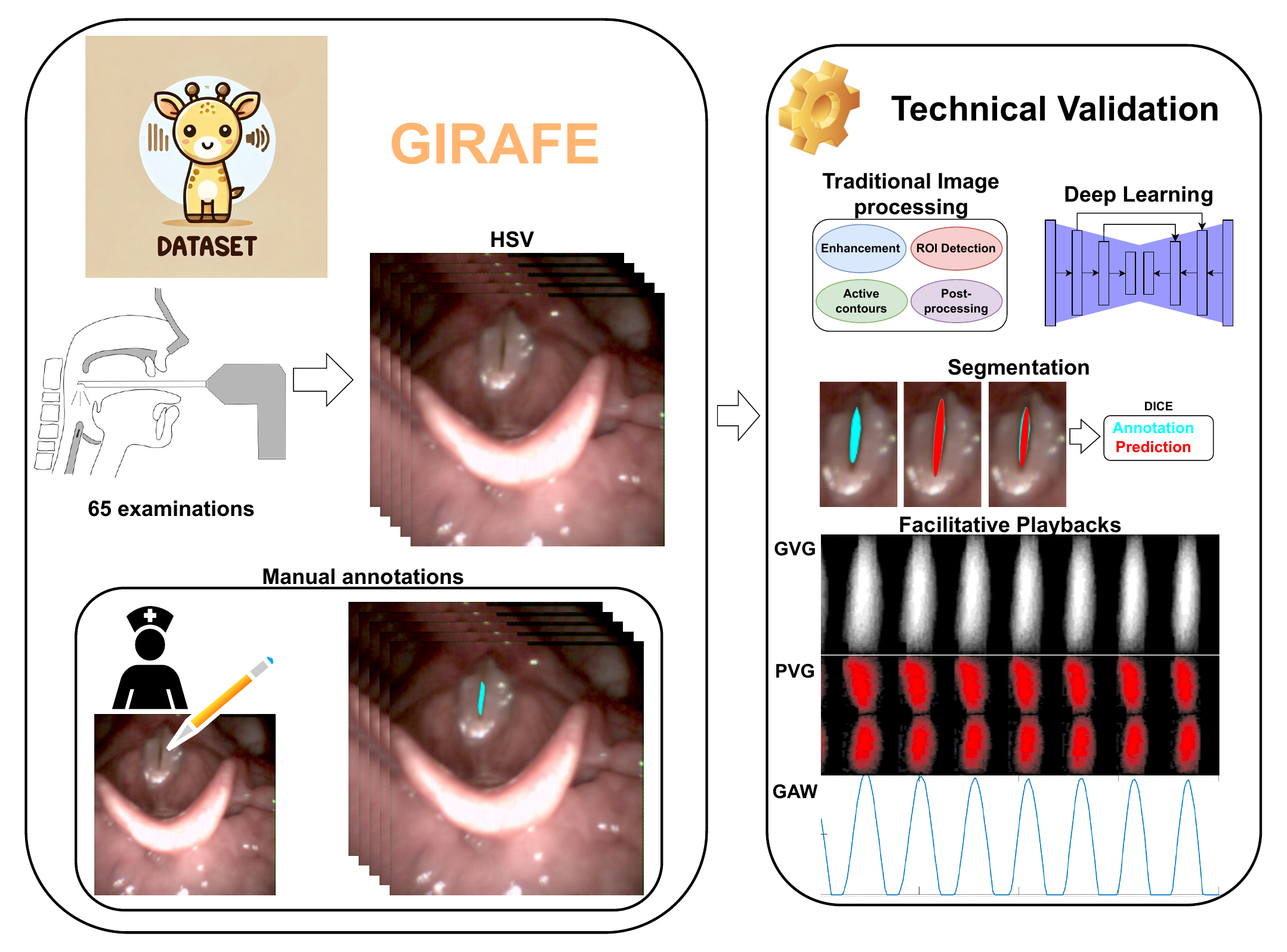}
\caption{Workflow for generating the GIRAFE dataset. Participants are of varying ages, gender, and health conditions, and were recruited at the Otorhinolaryngology Service of Hospital General  Gregorio Marañón in Madrid. Glottal segmentation was conducted using manual, automatic, and semi-automatic techniques. To validate the GIRAFE dataset, two deep neural networks were trained, and the resulting segmentations were found to be highly consistent with the expert manual annotations.}
\label{fig:datasetWorkflow}
\end{figure}


\section{Data Description}

GIRAFE aims to contribute with open data but also with open software to provide a robust and reliable baseline for the semantic segmentation of the glottal gap and for the evaluation of FP. To achieve this, it includes a diverse set of recordings collected at the facilities of the Otorhinolaryngology Service of the Gregorio Marañon Hospital in Madrid, which were carefully delineated with their corresponding manual annotation masks. To validate the benchmark data and establish a baseline score, GIRAFE includes segmentation results obtained using two well-established image processing techniques: \textbf{\texttt{InP}} \cite{Andrade-Miranda2017Inpa} and \textbf{\texttt{Loh}} \cite{Lohscheller2007}. Additionally, to keep pace with advancements in the field, two state-of-the-art DL methods were used in this paper, namely: UNet \cite{RFB15a} and SwinUnetV2 \cite{liu2022swin}, both trained using the GIRAFE dataset. 

The repository contains raw HSV recordings in AVI format for each patient, accompanied by metadata in JSON format. The glottal segmentation and FP results from all approaches, including the manual segmentation, are saved in PKL Phyton \textsuperscript{®} format. The dataset also includes 760 images in PNG format with the corresponding manual annotations, along with the default splitting used for all DL models in JSON format. The complete workflow of GIRAFE, encompassing data collection, annotation, and validation processes, is shown in Fig.\ref{fig:datasetWorkflow}. 

This section is organized as follows: First, we outline the data collection strategy, specifying the number of recordings and the population involved. Next, we describe the data acquisition process and its characteristics, including the devices used and the clinical findings. We then explain the data labelling and evaluation methods, including the segmentation techniques and the use of FP. Finally, we address the ethical considerations and relevant declarations.

\subsection{Data collection.}

To achieve high temporal resolution and accurately capture the fast motion of the vocal folds, recordings were performed using HSV. This technique allows the diagnosis of subtle vibratory patterns and anomalies undetectable with techniques based on standard laryngeal videostroboscopy (VS) \cite{Deliyski2008,kendall2010laryngealCH14}.

The population selected to create GIRAFE was meticulously chosen so the dataset includes recordings showing intricate details of the vocal folds motion, which is crucial for understanding various phonatory mechanisms and pathologies. By capturing the true intra-cycle vibratory behaviour of the vocal folds, the dataset visualizes aperiodic movements and transient phonatory events, such as phonatory breaks, laryngeal spasms, and the onset and offset of phonation \cite{Braunschweig2008,Unger2015,Herbst2014}. 

\begin{figure}[ht]
\centering
\includegraphics[width=1\linewidth]{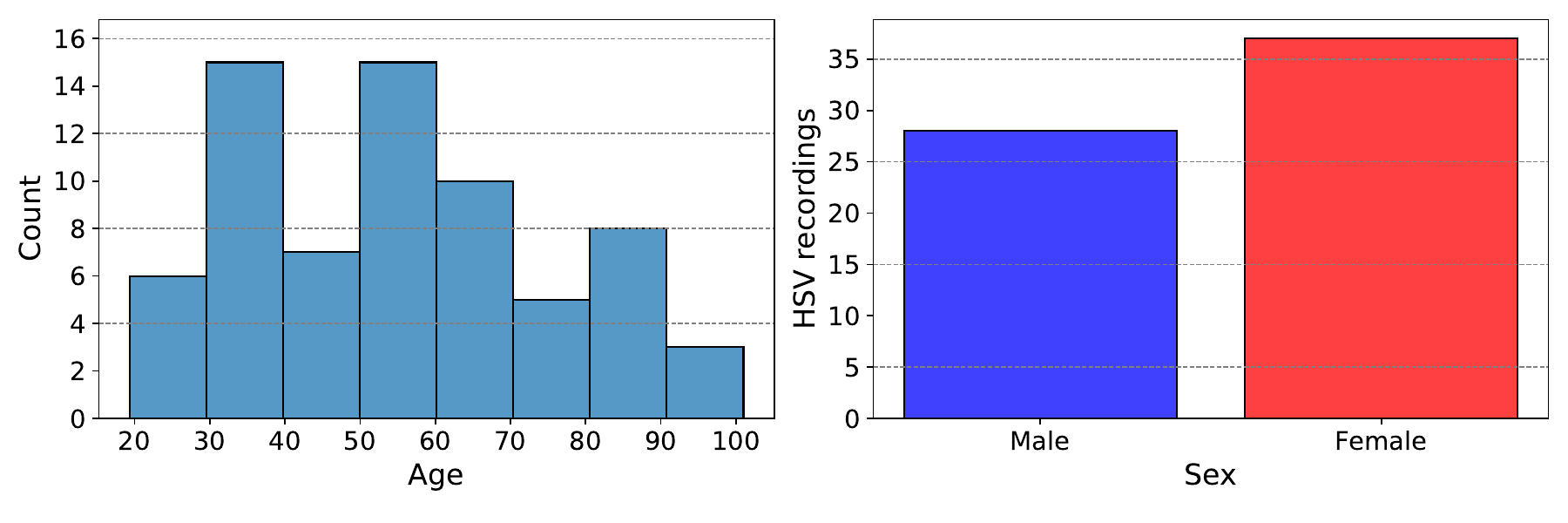}
\caption{Age and sex distribution of the recordings in the GIRAFE dataset.}
\label{fig:histogram}
\end{figure}

The HSV recordings were carried out by a specialist of the Otorhinolaryngology Service of Gregorio Marañón Hospital. The database comprises 65 recordings from 50 patients: 30 females (60\%) and 20 males (40\%), with a mean age of 55.65 ± 19.35 years, ranging from 25 to 101. The recordings capture a wide range of vocal fold behaviours, from normal function to various pathologies, providing a valuable resource for studying the biomechanics of voice production and disorders. Fig. \ref{fig:histogram} shows the age and sex distribution of the data, highlighting the diverse demographics of the study population. This diversity enhances the dataset's applicability, making it a robust tool for training deep-learning segmentation models. All patients were native Spanish speakers and adhered to the same experimental protocol.

\subsection{Data Acquisition and properties.}
The HSV sequences were acquired using the WOLF\textsuperscript{®} HRES ENDOCAM 5562 camera system and a rigid endoscope with a 70-degree angle of view. The light source was the WOLF AUTO LP 5132. The recordings exhibit varying levels of illumination, contrast, partial occlusion of the glottis, and lateral displacements of the camera, providing a comprehensive resource for robust analysis under diverse conditions. Some key features of the recordings are highlighted as follows:

\begin{itemize}
\item Each sequence contains 502 frames, resulting in 32,630 images available for analysis. This extensive frame count enables thorough temporal analysis, the study of dynamic changes over time, and the evaluation of different segmentation models.
\item The videos capture a sustained vowel phonation, including, in some cases, the vocal onset, providing valuable data on phonatory behavior. This aspect is particularly important for studying the biomechanics of voice production and the initiation of phonation, which are critical for diagnosing voice disorders.
\item The sampling rate was 4,000 fps, and the spatial resolution 256 $\times$ 256 pixels. Such a high frame rate is necessary to accurately capture the fast dynamics of vocal fold motion, while spatial resolution provides sufficient detail for identifying anatomical landmarks and pathologies.
\item The distance between the camera head in the oropharynx and the vocal folds varies, reflecting real-world clinical conditions. This variability enhances the dataset's robustness for developing and testing semantic segmentation algorithms.
\item All sequences were recorded in color, enhancing the visibility of anatomical features and pathologies. Color recordings facilitate better differentiation of tissue types and more accurate identification of pathological changes.
\end{itemize}

\begin{figure}[ht]
    \centering 
    \begin{subfigure}[b]{0.4\columnwidth}
        \centering 
        \includegraphics[width=0.85\columnwidth]{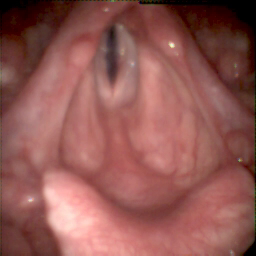}
        \caption{} 
        \label{fig:decoder_a}
    \end{subfigure} \hspace{-0.1cm}
    \begin{subfigure}[b]{0.4\columnwidth}
    \centering      
    \includegraphics[width=0.85\columnwidth]{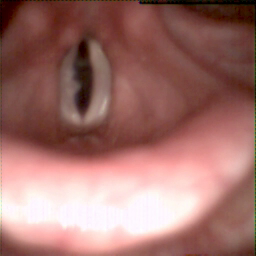}
    \caption{}
    \label{fig:decoder_b}
    \end{subfigure} \hspace{-0.1cm}
        \begin{subfigure}[b]{0.4\columnwidth}
    \centering      
    \includegraphics[width=0.85\columnwidth]{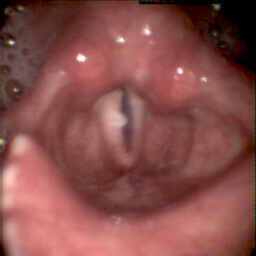}
    \caption{}
    \label{fig:decoder_c}
    \end{subfigure} \hspace{-0.1cm}
        \begin{subfigure}[b]{0.4\columnwidth}
    \centering      
    \includegraphics[width=0.85\columnwidth]{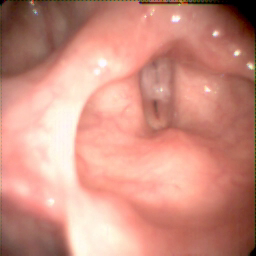}
    \caption{}
    \label{fig:decoder_d}
    \end{subfigure}
    \caption{\label{fig:disorders} Illustration of four cases with voice disorders: (a) cervicotomy for cervical hernia with postoperative diplophonia; (b) multinodular goiter; (c) vagal paraganglioma with fat infiltration; and, (d) vocal fold polyp.} \vspace{-0.4cm}
\end{figure}

The database includes voiced sounds in laryngeal mechanism M1 \cite{henrichbernardoni2022}, along with a variety of vocal fold pathologies. Figure \ref{fig:disorders} depicts examples of voice disorders included in the corpus, and Table \ref{tab:summary_pathology} summarizes the diagnosed voice disorders at the time of recording. The dataset totals 15 recordings of healthy subjects and 26 cases with identified disorders and/or noticeably affected vocal folds oscillations. Health status information was unavailable for 24 subjects.   

\begin{table}[]
\centering
\resizebox{0.8\textwidth}{!}{%
\begin{tabular}{|l|l|l|l|}
\hline
\rowcolor[HTML]{FFCE93} 
{\color[HTML]{000000} Disorder Status} & {\color[HTML]{000000} \# of videos} & {\color[HTML]{000000} Disorder Status} & {\color[HTML]{000000} \# of videos} \\ \hline
Healthy & 15 & Nodules & 2  \\ \hline
Multinodular Goiter & 1 & Polyps & 3 \\ \hline
Diplophonia & 2 & Cysts & 2 \\ \hline
Carcinoma & 2 & Other & 2\\ \hline
Paralysis & 7 & Unknown status & 24 \\ \hline
Paresis & 5 & &  \\ \hline
\end{tabular}%
}
\caption{Summary of voice disorders presented in the GIRAFE dataset.}
\label{tab:summary_pathology}
\end{table}

\subsection{Data labelling, segmentation, and facilitative playbacks}

An expert in glottis segmentation performed the manual delineations for the dataset, which were subsequently verified by an otolaryngologist. Manual segmentation was conducted on 38 out of the 65 videos, encompassing a complete glottal cycle per patient and resulting in 760 segmented frames. These segmentations were further validated using two established image processing techniques: \textbf{\texttt{InP}} \cite{Andrade-Miranda2017Inpa} and \textbf{\texttt{Loh}} \cite{Lohscheller2007}. 

The detailed segmentation process and FP generation are described as follows:

\begin{itemize}
    \item The \textbf{\texttt{InP}} method \cite{Andrade-Miranda2017Inpa} proposed an automatic segmentation method based on two steps.  First, the Region of Interest (ROI) (i.e., a broad area including the glottal gap) is automatically detected by evaluating the interframe intensity variation. Second, the glottis is delineated by combining background subtraction, inpainting, and a technique based on active contours. We applied this procedure to segment the complete dataset automatically. However, in cases where no intensity variation was observed, such as in paresis or partial paralysis (27 out of 65 cases), the first step failed. In these instances, the ROI was manually delineated and then proceeded with the segmentation. 
    \item The \textbf{\texttt{Loh}} method is a semiautomatic approach performed using the GlottalImageExplorer package \cite{Birkholz2016}, which is based on \cite{Lohscheller2007}. \textbf{\texttt{Loh}} was written in C++ using the cross-platform GUI library wxWidgets 2.8.12. Segmentation was performed on 39 out of the 65 HSV videos, and the time to segment 400 high-speed images was approximately 3.1 seconds, excluding the previous user interaction. The user-adjusted values of the thresholds averaged 53.4 $\pm$ 16.9 seconds.
    \item UNet and SwinUnetV2 architectures were also trained from scratch using the 760 manual annotations. These DL architectures are widely recognized for their accuracy and efficiency in medical image segmentation tasks.
    \item Three different FP were calculated: GAW \cite{Andrade-Miranda2020}, GVG \cite{Karakozoglou2012}, and PVG \cite{Lohscheller2008b} (see Fig.\ref{fig:datasetWorkflow}). GAW is a 1D representation of the glottal area as a function of time. PVG and GVG playbacks are 2D plots of vocal folds vibratory patterns as a function of time, summarizing glottal-edge movements along the posterior-anterior part into a time-varying image line. These representations have been shown to facilitate visual perception and increase the reliability of visual ratings while preserving the most relevant characteristics of glottal vibratory patterns.
    
\end{itemize}

\subsection{Data Organization}

\begin{figure}[ht!]
\includegraphics[width=1\linewidth]{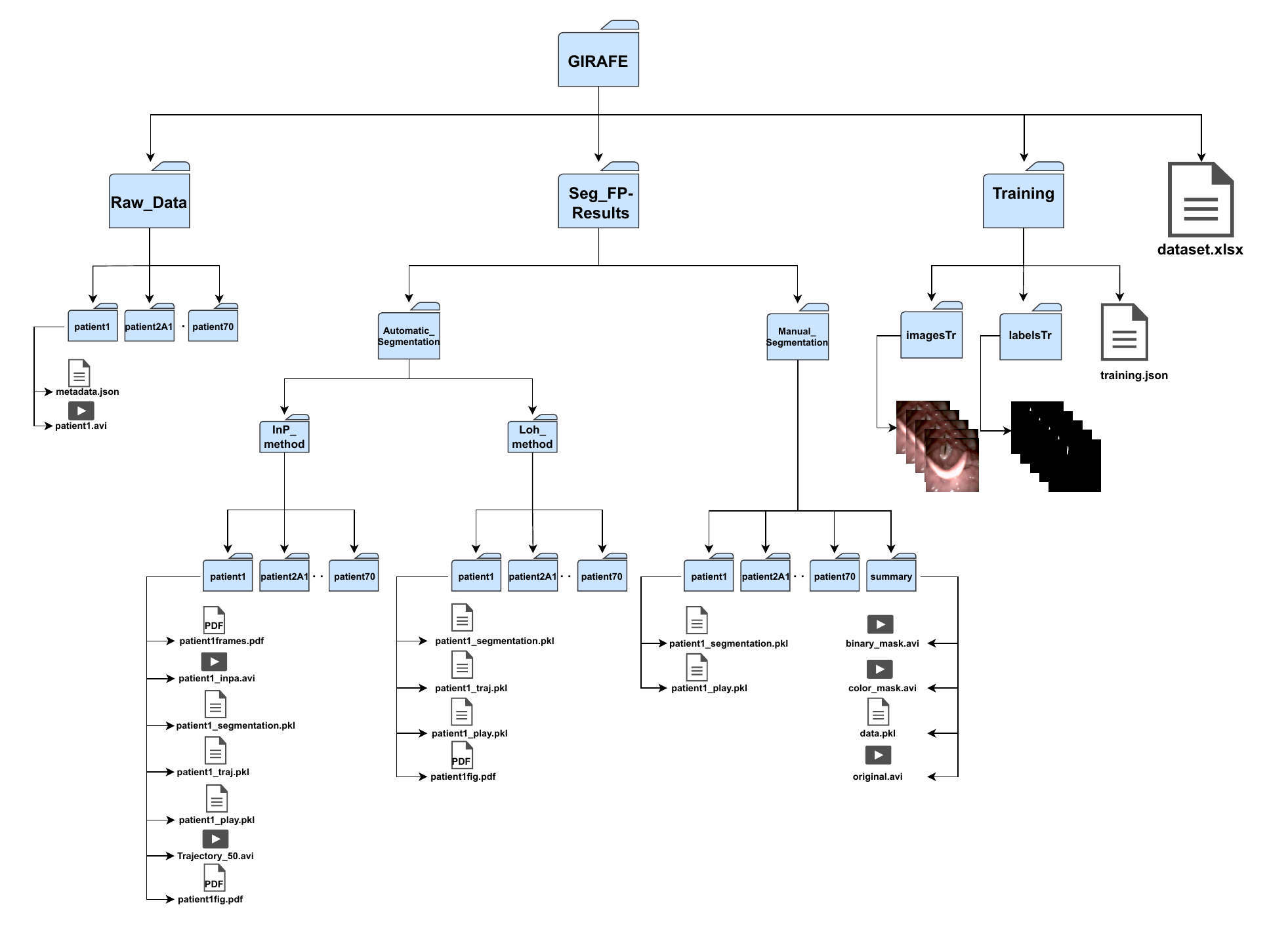}
\caption{Database tree structure: hierarchical representation of the data records and their organization within the database. For simplicity, the contents of only one patient are displayed for each subfolder.}
\label{fig:tree_folders}
\end{figure}

The structural organization of the dataset is shown in Fig. \ref{fig:tree_folders}. It comprises 32.8 GB of data, including 65 HSV recordings with their respective metadata and results obtained using traditional image processing methods.

The dataset is divided into three directories: \textbf{\texttt{\char`\\Raw\_Data}}, \textbf{\texttt{\char`\\Seg\_FP-Results}}, and \textbf{\texttt{\char`\\Training}}. The following sections offer a detailed overview of the contents of each directory.

\subsection{The \textbf{\texttt{\char`\\Raw\_Data}} directory}

The \textbf{\texttt{\char`\\Raw\_Data}} directory contains 65 folders (one per exploration), each containing an HSV recording. These folders are labeled with a unique \texttt{ID} in the format \textbf{\texttt{\char`\\Raw\_Data\char`\\patientID}}. Each folder contains the individual patient's recordings. This directory contains several patients with multiple recordings, taken on the same day or on different dates. For example, patient 51 has two recordings from the same day (stored in \textbf{\texttt{\char`\\Raw\_Data\char`\\patient51A1}} and \textbf{\texttt{\char`\\Raw\_Data\char`\\patient51A2}}) and one from a later date (saved in \textbf{\texttt{\char`\\Raw\_Data\char`\\patient51B1}}). 

Each \textbf{\texttt{\char`\\Raw\_Data\char`\\patientID}} folder contains two different files: 
\begin{itemize}
    \item \texttt{patientID.avi}. This file contains a video file in uncompressed AVI format, which includes the original laryngeal HSV recording.
    \item \texttt{metadata.json}. This file includes information about the patient, such as ID, age, sex, and disorder, as well as acquisition details, like camera type and video resolution. Furthermore, it specifies whether the recording is complemented with the manual segmentation (\textbf{\texttt{Man}}) and/or with one of the aforementioned automatic or semi-automatic segmentation techniques (\textbf{\texttt{InP}} or \textbf{\texttt{Loh}}). If \textbf{\texttt{Man}} or \textbf{\texttt{Loh}} is present, it is marked as "Yes"; otherwise, with "No". Additionally, all recordings are complemented, at least, with \textbf{\texttt{InP}} segmentations, calculated following a fully automatic or a semi-automatic procedure (labeled as "A" or "SA", respectively). Table \ref{tab:summary_metadata} exemplifies the information contained in the metadata file of a specific patient.
\end{itemize}

\begin{table}[h]
\centering
\resizebox{0.5\textwidth}{!}{%
\begin{tabular}{|l|l|}
\hline
\rowcolor[HTML]{FFCE93} 
{\color[HTML]{000000} Key} & {\color[HTML]{000000} Value} \\ \hline
ID &  1 \\ \hline
camera & WOLF HRES ENDOCAM 5562 \\ \hline
width &  256 \\ \hline
height &  256 \\ \hline
nb\_frames &  502 \\ \hline
age &  36 \\ \hline
subject sex &  Male  \\ \hline
recorded date &  10/04/2013  \\ \hline
disorder status &  Healthy  \\ \hline
InP &  A  \\ \hline
Loh &  Yes  \\ \hline
Man &  Yes  \\ \hline
\end{tabular}%
}
\caption{Example of a metadata JSON file from a specific HSV recording (\texttt{\char`\\Raw\_Data\char`\\patient1\char`\\metadata.json}).}
\label{tab:summary_metadata}
\end{table}

\subsection{The \textbf{\texttt{\char`\\Seg\_FP-Results}} directory}
\begin{figure}[ht!]
\includegraphics[width=1\linewidth,trim={0cm 0.3cm 0 0.5cm},clip]{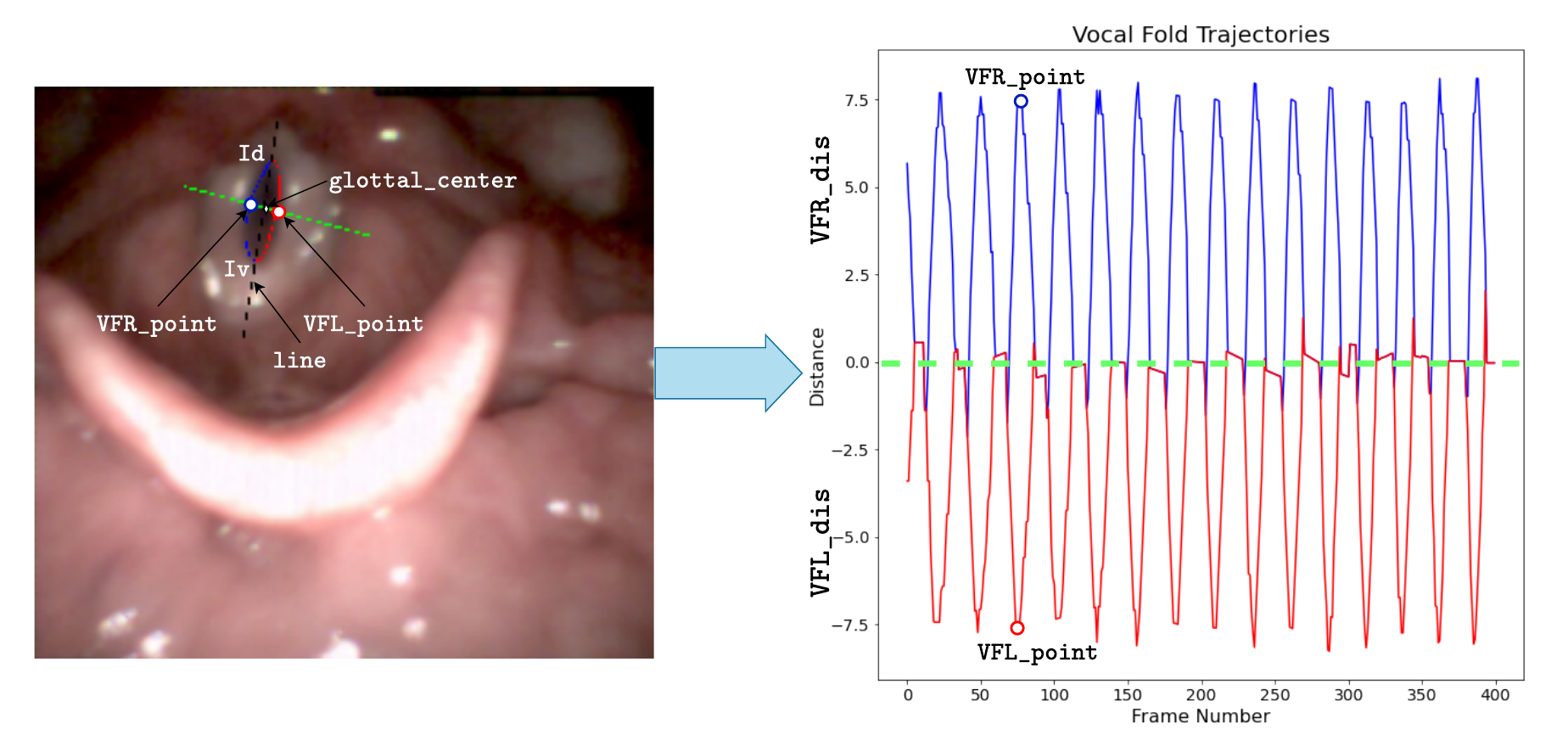}
\caption{Visual illustration of the dictionary keys in \texttt{patientID\_play.pkl} and \texttt{patientID\_traj.pkl} of an example in folder \textbf{\texttt{\char`\\Seg\_FP-Results\char`\\Automatic\_Segmentation}}. The plot on the right hand side represents the trajectories (i.e., the distance travelled along the main orthogonal axis represented in green) of the left (in red) and right vocal folds (in blue).}
\label{fig:variables}
\end{figure}

The \textbf{\texttt{\char`\\Seg\_FP-Results}} directory contains 2 folders. They are enumerated and  presented next:
\begin{itemize}
\item The folder \textbf{\texttt{\char`\\Seg\_FP-Results\char`\\Automatic\_Segmentation}} contains the segmentation results of the aforementioned automatic and semi-automatic methods. In turn, it is organized around two subfolders, which are listed below. 

The subfolder \textbf{\texttt{\char`\\Seg\_FP-Results\char`\\Automatic\_Segmentation\char`\\Loh\_method}}  contains 39 additional folders, each labeled as \textbf{\texttt{\char`\\Seg\_FP-Results\char`\\Automatic\_Segmentation\char`\\Loh\_method\char`\\patientID}}. Each of these subfolders contains three files in PKL format and one PDF that includes the segmentation results \cite{Lohscheller2007}, trajectories, and FP. They are enumerated next, along with their specific content: 
    \begin{itemize}
        \item \texttt{patientID\_segmentation.pkl}. This file contains the segmentation results stored in a Phyton dictionary with the following fields: 
                \begin{itemize}
                \setlength\itemsep{-0.15em}
                \item \texttt{"mask":} NumPy array containing the segmentation (1 for the foreground vocal fold, 0 for the background).
                \item \texttt{"contours":} NumPy array containing the segmentation contours (1 for the foreground vocal fold, 0 for the background).
                \item \texttt{"contours\_color":} NumPy array containing the segmentation contours in color (left fold represented in blue, right fold represented in red) overlaid on the original video.
                \item \texttt{"edges":} This is a Python dictionary containing the following keys (see Fig. \ref{fig:variables} for a visual representation of them): 
                        \begin{itemize}
                        \setlength\itemsep{-0.1em}
                        \item \texttt{"Ri":} Regression line that splits the left and right folds.
                        \item \texttt{"left":} Points representing the left vocal fold.
                        \item \texttt{"right":} Points representing the right vocal fold.
                        \item \texttt{"Iv":} Ventral points.
                        \item \texttt{"Id":} Dorsal points.
                        \end{itemize}
                \end{itemize}
        \item \texttt{patientID\_traj.pkl}. This file contains an analysis of the the vocal folds trajectory referenced to the center of the glottal axis's. This is a Phyton dictionary, which includes the following fields (see Fig. \ref{fig:variables} for a visual interpretation of them): 
                \begin{itemize}
                \setlength\itemsep{-0.15em}
                \item \texttt{"VFL\_dis":} Distance between the glottal axis and the left vocal fold.
                \item \texttt{"VFR\_dis":} Distance between the glottal axis and the right vocal fold.
                \item \texttt{"VFL\_Point":} Left intersection point between the glottal axis and a line perpendicular to the center of the glottal axis.
                \item \texttt{"VFR\_Point":} Right intersection point between the glottal axis and a line perpendicular to the center of the glottal axis.
                \item \texttt{"glottal\_center":} Center point of the glottal axis. 
                \item \texttt{"line":} Equation representing the glottal axis.
                \end{itemize}

        \item \texttt{patientID\_play.pkl}. This is a Phyton dictionary containing the FP synthesized. It is organized around the following fields (see Fig. \ref{fig:variables} for a visual representation of them): 
                \begin{itemize}
                \setlength\itemsep{-0.15em}
                \item \texttt{"VFL\_dis":} Distance between the glottal axis and the left vocal fold.
                \item \texttt{"VFR\_dis":} Distance between the glottal axis and the right vocal fold.
                \item \texttt{"VFL\_Point":} Left intersection point between the glottal axis and a line perpendicular to the center of the glottal axis.
                \item \texttt{"VFR\_Point":} Right intersection point between the glottal axis and a line perpendicular to the center of the glottal axis.
                \item \texttt{"glottal\_center":} Center point of the glottal axis. 
                \item \texttt{"line":} Equation representing the glottal axis.
                \end{itemize}

        \item \texttt{patientIDfig.pdf}. This PDF file illustrates four different FP synthesized taking the segmentations as an input: GVG, PVG, DKG, and GAW. 
        \end{itemize}

The subfolder \textbf{\texttt{\char`\\Seg\_FP-Results\char`\\Automatic\_Segmentation\char`\\InP\_method}}  contains 65 additional folders each named with \textbf{\texttt{\char`\\Seg\_FP-Results\char`\\Automatic\_Segmentation\char`\\InP\_method\char`\\patientID}}. These folders store the results of the segmentation according to the \textbf{\texttt{InP}} method,  including the same four files as the \textbf{\texttt{Loh}} method, along with three additional visualization files detailed below:
\vspace{-2mm}
    \begin{itemize}
    \item \texttt{patientIDframes.pdf}. This PDF file contains an illustration of the segmented glottal gap, covering part of the glottal cycle. This visualization is generated from the \texttt{patientID\_segmentation.pkl} file. 
    \item \texttt{patientID\_inpa.avi}. This is a video in AVI format which displays the segmentation results in binary format. 
    \item \texttt{trajectory\_50.avi}. This is a video file in AVI format displaying the procedure for computing vocal fold trajectories based on the \texttt{patientID\_traj.pkl} and \texttt{patientID\_segmentation.pkl} files.  
    \end{itemize}

\item Folder 
\textbf{\texttt{\char`\\Seg\_FP-Results\char`\\Manual\_Segmentation}}: 

This folder contains 38 subfolders, again labeled with \textbf{\texttt{\char`\\Seg\_FP-Results\char`\\Manual\_Segmentation\char`\\patientID}}, each containing the manual segmentations. Each subfolder contains two files in PKL format: 
\begin{itemize}
    \item \texttt{patientID\_play.pkl} 
    \item \texttt{patientID\_segmentation.pkl}. Unlike the previously described \texttt{patientID\_segmentation.pkl}, this file includes only two entries: one for the original video (\texttt{original}) and the other for the binary segmentation (\texttt{mask}), both stored as NumPy arrays. 

\end{itemize}

An additional subfolder, \textbf{\texttt{\char`\\Seg\_FP-Results\char`\\Manual\_Segmentation\char`\\Summary}}, contains a visual representation of all manually segmented frames, also including the binary masks. This folder contains the following files: 
\begin{itemize}
    \item \texttt{binary\_mask.avi}. This is a video sequence in AVI format, which displays the motion of vocal folds in a binary format. 
    \item \texttt{color\_mask.avi}. This is an AVI video sequence that displays the movement of the vocal folds in color.
    \item \texttt{original.avi}. This is a video file in AVI format that shows a motion with the subset of original frames used for the segmentation process. 
    \item \texttt{data.pkl}. This is a PKL file that contains the masks in a Python dictionary file.
\end{itemize}

\end{itemize}

\subsection{The \textbf{\texttt{\char`\\Training}} directory.}

The \textbf{\texttt{\char`\\Training}} directory is organized to facilitate the training and testing of DL semantic segmentation models. It is organized into two folders: \textbf{\texttt{\char`\\Training\char`\\imagesTr}}, containing the original frames; and \textbf{\texttt{\char`\\Training\char`\\labelsTr}}, which contains the corresponding manual segmentations. Each subfolder contains 760 images in uncompressed PNG format, corresponding to the same number of consecutive frames extracted from the original sequences. 

Additionally, this directory also includes one file, which is described next: 

\begin{itemize}
    \item \texttt{training.json}. This is a JSON file containing the distribution of patients to be used for the training and testing sets used to evaluate the performance of the DL models included in the repository (see usage notes).
\end{itemize}

\subsection{Clinical Data and Recording Details}
Clinical data are stored in a spreadsheet file in XLSX format named \texttt{dataset.xlsx}. This file is placed in the root of the hierarchical structure of the corpus, and provides a summary of the clinical data collected for each patient, including details such as disease status, sex, date of birth, age, and recording date. Additionally, it includes information on the availability of contours delineated using manual (\textbf{\texttt{Man}}), automatic (\textbf{\texttt{Loh}}) and semiautomatic techniques (\textbf{\texttt{InP}}). This file is included to provide an initial, comprehensive overview of the dataset, enabling a clearer understanding of its structure and contents. Additional metadata, including camera type and resolution, are available for each video in the corresponding \texttt{metadata.json} files, located in the \textbf{\texttt{\char`\\Raw\_Data}} folder.

\section*{Experimental design, Material and Methods}

Various subsets of the GIRAFE corpus have been extensively used in the past as an experimental basis for several research works \cite{Andrade-Miranda2015, Andrade-Miranda2017Inpa, andrade2017analyzing, Andrade-Miranda2020, ANDRADEMIRANDA2017}. 

Specifically, the dataset was widely used to evaluate the accuracy of different glottal segmentation methods and their suitability to synthesize the motion of the vocal folds using different FP \cite{Andrade-Miranda2017Inpa}. The corpus was first used to objectively evaluate and compare traditional image processing techniques, such as background subtraction and active contours (\textbf{\texttt{InP}} \cite{Andrade-Miranda2017Inpa}), region growing techniques (\textbf{\texttt{Loh}} \cite{Lohscheller2007}), and watershed transform \cite{Andrade-Miranda2015}. In the same work, a subjective evaluation was also carried out using the information provided by different FP to detect segmentation errors by analyzing the GVG, PVG, and GAW \cite{Andrade-Miranda2017Inpa}. Specifically, the assessment aimed to identify discrepancies in vibratory patterns, segmentation errors in the anterior or posterior glottis, playback discontinuities, and inconsistencies in the duration of the open phase. The experiments demonstrated objective and subjective improvements in segmentation quality, readability, and shape similarity of the FP using the \textbf{\texttt{InP}} method. 

In addition to the works mentioned above, an alternative validation was included in this work using two DL methods. This validation is not intended to be exhaustive but aims to provide new evidence of the corpus's usefulness and demonstrate an example of how to use the dataset (see the \textit{Usage Notes} section). In this regard, a comparison is provided between the two aforementioned methods based on traditional image processing (\textbf{\texttt{InP}} and \textbf{\texttt{Loh}}) and two approaches based on DL (UNet and SwinUnetV2). 

\begin{figure}[!ht]
\centering
\includegraphics[width=1\linewidth,trim={2cm 3.5cm 2cm 3.5cm},clip]{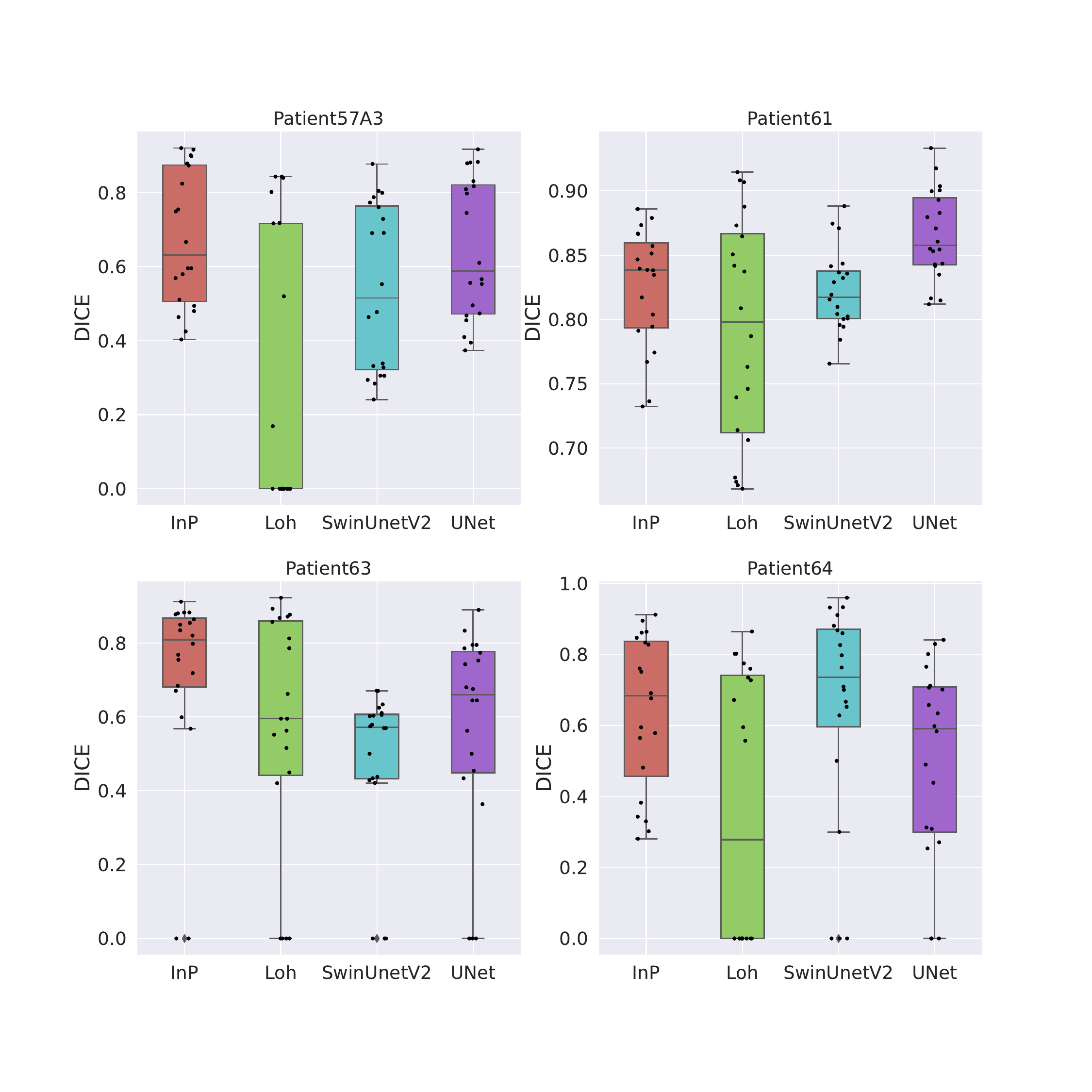}
\caption{Boxplots illustrating the DICE metric for the four test patients.}
\label{fig:Boxplot}
\end{figure}

\begin{figure*}[!ht]
\raggedleft
\resizebox{0.85\textwidth}{!}{%
\input{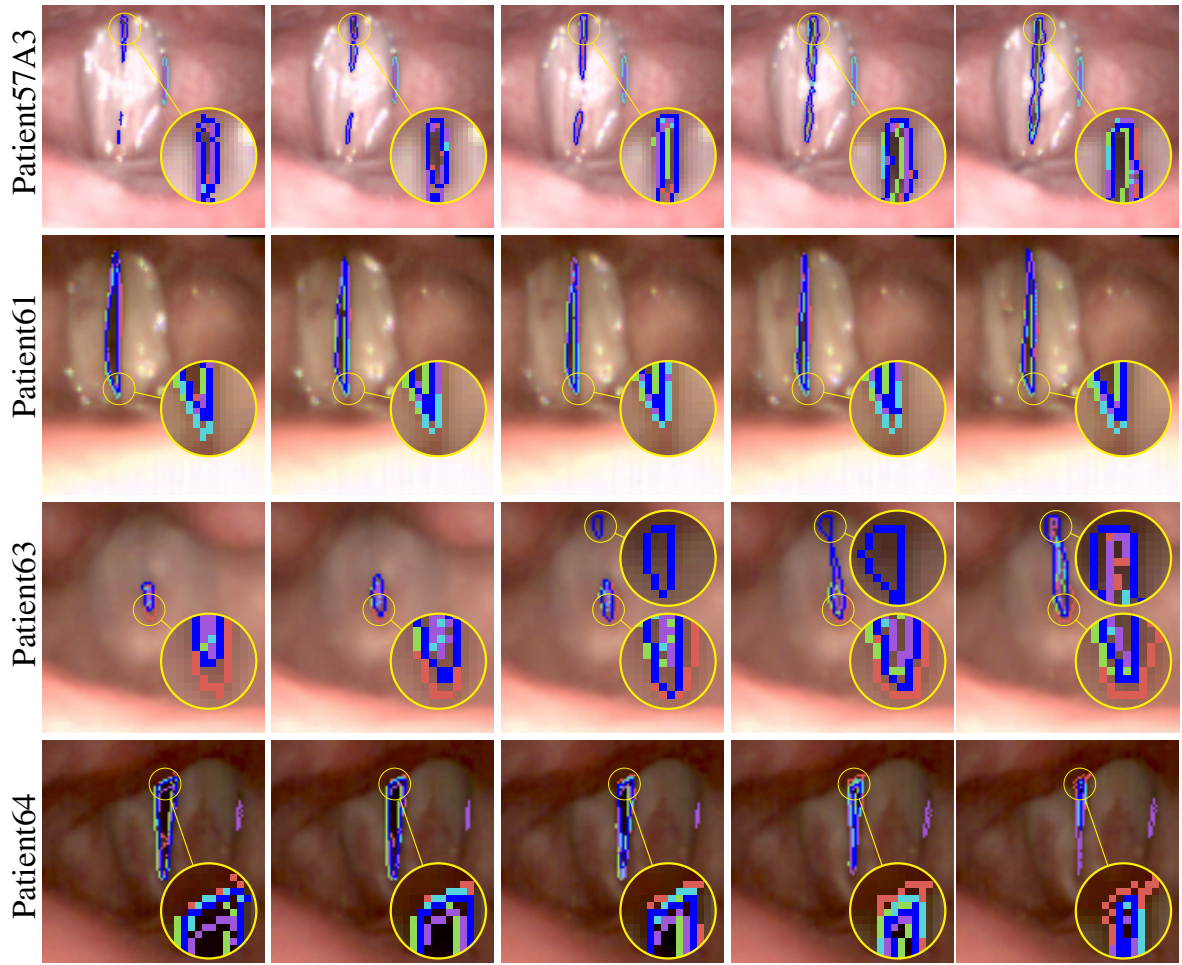}
}
\caption{Visual representation of segmentation results, with blue contours indicating ground truth \statsquare{Man}. The models are shown as \textbf{\texttt{InP}} \statsquare{InP}, \textbf{\texttt{Loh}} \statsquare{Loh}, UNet \statsquare{Unet}, and SwinUnetV2 \statsquare{Swin}, with their corresponding color codes in parentheses. Each row represents a different patient, displaying five consecutive frames.}
\label{fig:visualImage}
\end{figure*}

To ensure a fair and consistent comparison between the DL models, identical preprocessing steps were applied. Images were resized to $256 \times 256$, and segmentation was performed in RGB for both models. In line with BAGLS \cite{Gomez2020}, data augmentation techniques were applied on the fly during training, including random rotation, scaling, flipping, Gaussian noise, Gaussian blur, and brightness and contrast adjustments using the MONAI Python package \cite{cardoso2022monai}. Both models were trained for 200 epochs with a batch size of 16, utilizing the Adam optimizer and the Dice loss function. The learning rate was set to $2 \times 10^{-4}$, with a weight decay of $1 \times 10^{-5}$. Out of the 38 manually segmented recordings, 30 were used for training, 4 for validation, and 4 for testing, each consisting of 20 frames. Table \ref{tab:resultsModels} summarizes the performance metrics, including the DICE score, Jaccard index, Recall, and Precision, for the test patients across the four segmentation methods. A visual comparison of the DICE score distribution for each patient is shown in Fig. \ref{fig:Boxplot}, where boxplots illustrate the performance variation among the methods. Additionally, detailed visual results are provided in Fig. \ref{fig:visualImage}, showcasing the glottal gap segmentation for five consecutive frames corresponding to the same patients.  

%
\begin{table}[!h]
\centering
\resizebox{0.7\textwidth}{!}{
\begin{tabular}{c|c|c|c|c|c|}
\cline{2-6}
\rowcolor[HTML]{FFCE93}
                            & {\color[HTML]{000000}Method}     & {\color[HTML]{000000}DICE}           & {\color[HTML]{000000}Jaccard}           & {\color[HTML]{000000}Recall}         & {\color[HTML]{000000}Precision}      \\ \hline
\multicolumn{1}{|c|}{\multirow{4}{*}{Patient57A3}} & InP        & \textcolor{red}{0.675}                & \textcolor{red}{0.538}                & 0.567                                 & \textcolor{red}{0.946} \\ \cline{2-6} 
\multicolumn{1}{|c|}{}                             & Loh        & 0.273                                 & 0.221                                 & 0.24                                  & 0.347          \\ \cline{2-6} 
\multicolumn{1}{|c|}{}                             & UNet       & \textcolor{blue}{0.646}               & \textcolor{blue}{0.506}               & \textcolor{red}{0.789}                & \textcolor{blue}{0.586}          \\ \cline{2-6} 
\multicolumn{1}{|c|}{}                             & SwinUnetV2 & 0.542                                 & 0.404                                 & \textcolor{blue}{0.769}               & 0.468          \\ \hline
\multicolumn{1}{|c|}{\multirow{4}{*}{Patient61}}   & InP        & \textcolor{blue}{0.825}               & \textcolor{blue}{0.704}               & \textcolor{red}{0.965}                & 0.722          \\ \cline{2-6} 
\multicolumn{1}{|c|}{}                             & Loh        & 0.792                                 & 0.664                                 & 0.792                                 & 0.798          \\ \cline{2-6} 
\multicolumn{1}{|c|}{}                             & UNet       & \textcolor{red}{0.866}                & \textcolor{red}{0.765}                & \textcolor{blue}{0.923}               & \textcolor{red}{0.824} \\ \cline{2-6} 
\multicolumn{1}{|c|}{}                             & SwinUnetV2 & 0.822                                  & 0.699                                & 0.843                                 & \textcolor{blue}{0.814}          \\ \hline
\multicolumn{1}{|c|}{\multirow{4}{*}{Patient63}}   & InP        & \textcolor{red}{0.711}                & \textcolor{red}{0.597}                & \textcolor{red}{0.714}                & 0.721          \\ \cline{2-6} 
\multicolumn{1}{|c|}{}                             & Loh        & 0.562                                 & \textcolor{blue}{0.454}               & \textcolor{blue}{0.492}               & 0.703          \\ \cline{2-6} 
\multicolumn{1}{|c|}{}                             & UNet       & \textcolor{blue}{0.566}               & 0.44                                  & 0.453                                 & \textcolor{blue}{0.823}          \\ \cline{2-6} 
\multicolumn{1}{|c|}{}                             & SwinUnetV2 & 0.477                                 & 0.335                                 & 0.337                                 & \textcolor{red}{0.84}  \\ \hline
\multicolumn{1}{|c|}{\multirow{4}{*}{Patient64}}   & InP        & \textcolor{blue}{0.639}               & \textcolor{blue}{0.505}               & \textcolor{red}{0.724}                & \textcolor{blue}{0.604}          \\ \cline{2-6} 
\multicolumn{1}{|c|}{}                             & Loh        & 0.364                                 & 0.29                                  & 0.345                                 & 0.402          \\ \cline{2-6} 
\multicolumn{1}{|c|}{}                             & UNet       & 0.495                                 & 0.371                                 & 0.562                                 & 0.479          \\ \cline{2-6} 
\multicolumn{1}{|c|}{}                             & SwinUnetV2 & \textcolor{red}{0.644}                & \textcolor{red}{0.542}                &\textcolor{blue}{0.658}                & \textcolor{red}{0.692} \\ \hline
\end{tabular}
}
\caption{Evaluation of different segmentation methods using different performance metrics (DICE, Jaccard, Recall, and Precision) on the four test recordings. The best scores are in red, and the second-best in blue.}
\label{tab:resultsModels}
\end{table}

Although the results with \textbf{\texttt{InP}} demonstrate superior performance, both UNet and SwinUnetV2 exhibit consistent outcomes across all four test patients, comparable to those achieved with \textbf{\texttt{InP}}. As shown in Fig. \ref{fig:Boxplot}, for Patient57A3, \textbf{\texttt{InP}} achieves the best performance with an average DICE score of $0.675$, closely followed by UNet with a score of $0.646$. SwinUnetV2 performs slightly worse due to the presence of additional artifacts near the vocal folds, leading to over-segmentation issues, as illustrated in the first row of Fig. \ref{fig:visualImage}. In contrast, for Patient61, UNet achieves the highest performance, followed by \textbf{\texttt{InP}}, with SwinUnetV2 closely trailing. The corresponding average DICE scores are 0.866, 0.825, and 0.822, respectively. The segmentations produced by all three models closely align with the ground truth, as shown in the second row of Fig. \ref{fig:visualImage}. For Patient63, \textbf{\texttt{InP}} once again outperforms the DL models, achieving an average DICE score of $0.711$. This case is particularly challenging, as the glottal gap is split into two parts. As seen in the third row of Fig. \ref{fig:visualImage}, most models fail to accurately segment the dorsal region of the vocal folds. In the case of Patient64, the performance of all models is visually satisfactory, as shown in the fourth row of Fig. \ref{fig:visualImage}. Among them, SwinUnetV2 achieves the highest DICE score of $0.644$, outperforming the other models. UNet, on the other hand, exhibits a significant drop in performance, with a DICE score of $0.495$, likely due to over-segmentation issues similar to those observed in Patient57A3. Furthermore, as illustrated in the boxplot (Fig. \ref{fig:Boxplot}), some frames achieve a DICE score of $0$. This occurs when the segmentation models incorrectly predict a closed glottal gap, despite the gap being partially open, with a few visible pixels still present. 

 The previous results provide evidence of the interest of the corpus for the experimentation, as it encompasses a diverse range of pathologies and recording conditions, including varying levels of illumination, contrast, the presence of glottal chinks, and lateral displacements of the camera. This diversity ensures that the dataset is both challenging and representative, making it suitable for evaluating the robustness and generalization capabilities of segmentation models.


\subsection{Usage Notes}

As commented, the utility of the GIRAFE corpus has also been evaluated using two different state-of-the-art DL techniques, namely: UNet and SwinUnetV2. 

The corresponding Python data management scripts can be found in the GitHub repository referenced in the Code Availability section. For practical guidance and a hands-on demonstration, the corpus includes a Jupyter notebook named \texttt{Seg\_FP-Results.ipynb}, which delineates the process of reading images and their masks, as well as the process to follow in order to generate several FP visualizations. MATLAB scripts (\textbf{\texttt{\char`\\Matlab\_code}}) for generating FP from raw segmentations are also provided, with detailed instructions available in the \texttt{Matlab\_code.ipynb} notebook.

The repository also provides source code for two basic experiments using the referenced DL architectures, available in the \textbf{\texttt{\char`\\DL\_code}} folder, as detailed in the \textit{Technical Validation} section. This folder includes scripts for training (\texttt{train.py}) and testing (\texttt{inference.py}) the DL models, serving as a practical guide for working with the dataset and offering a foundation for further exploration and analysis with new methods. These experiments are not designed to maximize accuracy or improve state-of-the-art results but to demonstrate how the dataset can be used.

\section*{Ethics declaration}

The study was approved by the Ethics Review Board of the Hospital Universitario Gregorio Marañón in Madrid (code 11/2015), following the guidelines of the Spanish Ethical Review Act. All participants filled out a questionnaire and provided written consent to join the study. Patients were fully informed of their rights, including their ability to withdraw from the study at any time. All participants were native Spanish speakers and followed the same experimental protocol. The clinical data was collected by the otolaryngologist who performed the procedures and was the only person in direct contact with the patients. To maintain anonymity, each patient was assigned a unique ID, separated from their hospital clinical history code. No personal information was shared with the researchers who had access to the dataset. 

\section*{Data and Code availability}

The GIRAFE dataset is available through a Zenodo repository \cite{girafe_zenodo}.

The codes used for data processing, DL training and calculation of the FP are available on a dedicated GitHub\textsuperscript{®} repository (\url{https://github.com/Andrade-Miranda/GIRAFE}).

The \textbf{\texttt{InP}} method is publicly accessible via the GitHub repository (\url{https://github.com/Andrade-Miranda/Glottal-Gap-Segmentation}).

The \textbf{\texttt{Loh}} method, referred to as the GlottalImageExplorer toolbox \cite{Birkholz2016}, can be freely downloaded from \url{https://www.vocaltractlab.de/index.php?page=glottalimageexplorer-download}.

\section*{CRediT Author Statement}

G.A.M. and J.I.G. designed the experiment. G.A.M., K.C. and J.I.G. designed the methodology. G.A.M. and K.C. validated the experiment. D.P.S. collected the data. J.I.G. provided the resources. G.A.M. and J.I.G. wrote the initial draft version. G.A.M., J.D.A. and J.I.G. reviewed and edited the manuscript. G.A.M. and J.D.A. provided the software to analyse the data. J.D.A. and J.I.G. supervised. J.I.G. administered the project. J.I.G. acquired the funding. All authors have read and agreed to the published version of the manuscript.

\section*{Acknowledgements} 

This work was funded by the Ministry of Economy and Competitiveness of Spain (grants TEC-2012-38630-C04-01, DPI2017-83405-R1, PID2021-128469OB-I00 and TED2021-131688B-I00), and by Comunidad de Madrid, Spain. Universidad Politécnica de Madrid supports Julián D. Arias-Londoño through a María Zambrano UP2021-035 grant funded by European Union-NextGenerationEU.

The authors thank the Otorhinolaryngology Service of Hospital General Gregorio Marañón, Madrid, for the facilities provided.

The authors also thank the Madrid ELLIS unit (European Laboratory for Learning \& Intelligent Systems) unit for its indirect support, and to all patients who selflessly participated in the study.

Finally, the authors  acknowledge Universidad Politecnica de Madrid for providing computing resources on the Magerit Supercomputer.

\section*{Declaration of Competing interests} 

The authors declare that they have no competing interests.

\bibliographystyle{elsarticle-num} %
\bibliography{sample}


\end{document}